\documentclass{article}

     \usepackage[final,nonatbib]{neurips_2020}

\usepackage[utf8]{inputenc} 
\usepackage[T1]{fontenc}    
\usepackage{hyperref}       
\usepackage{url}            
\usepackage{booktabs}       
\usepackage{amsfonts}       
\usepackage{nicefrac}       
\usepackage{microtype}      
\usepackage{amsmath,amssymb,amsthm,amsfonts} 
\usepackage{color}          
\usepackage{graphicx}
\usepackage{bm}
\usepackage{setspace}
\def\fig#1{fig.~\ref{#1}}
\def\eq#1{eq.~\ref{#1}}
\def\tab#1{table~\ref{#1}}
\def\sec#1{section~\ref{#1}}

\title{Twin Neural Network Regression is a \\ Semi-Supervised Regression Algorithm}

\author{Sebastian J.~Wetzel\\
  Perimeter Institute for Theoretical Physics\\
  Waterloo, Ontario N2L 2Y5, Canada\\
  \texttt{swetzel@perimeterinstitute.ca}

  \And
  Roger G.~Melko\\

  Perimeter Institute for Theoretical Physics\\
  Waterloo, Ontario N2L 2Y5, Canada\\
  Department of Physics and Astronomy\\
  University of Waterloo\\
  Waterloo, Ontario, N2L 3G1, Canada \\
  \texttt{rgmelko@uwaterloo.ca}
  \And
  Isaac Tamblyn\\
  National Research Council of Canada\\
  Ottawa, Ontario K1N 5A2, Canada \\

  Vector Institute for Artificial Intelligence\\
  Toronto, Ontario M5G 1M1, Canada\\
  \texttt{isaac.tamblyn@nrc.ca}
}

\begin{document}

\maketitle

\begin{abstract}

Twin neural network regression (TNNR) is a semi-supervised regression algorithm, it can be trained on unlabelled data points as long as other, labelled anchor data points, are present. TNNR is trained to predict differences between the target values of two different data points rather than the targets themselves. By ensembling predicted differences between the targets of an unseen data point and all training data points, it is possible to obtain a very accurate prediction for the original regression problem. Since any loop of predicted differences should sum to zero, loops can be supplied to the training data, even if the data points themselves within loops are unlabelled. Semi-supervised training improves TNNR performance, which is already state of the art, significantly.

\end{abstract}
\section{Introduction}

Regression is the process of estimating the relationship between feature variables to outcome variables. It is one of the most central machine learning tasks in a wide range of scientific, analytic and industrial research and development disciplines. Regression analysis is considered to be a supervised machine learning task, where one is given labelled training data from which a model is trained to make predictions of the labels of unlabelled data points. In many applications, there is few labelled training data available, while unlabelled data is much easier to obtain. In addition to labelled data, unlabelled data can be leveraged to train machine learning models in the context of semi-supervised learning \cite{zhu2009introduction,chapelle2009semi}. There are two different cases of semi-supervised learning, which differ by the availability of the data for which one wants to obtain predictions during the training phase. The goal of inductive semi-supervised learning is to find the function that maps a feature variable to its outcome variable from a combined set of labelled and unlabelled data. This function can then be used to make predictions on new data points that are not available during the training phase. The goal of transductive semi-supervised learning is to infer the correct labels for the given unlabelled data, which is already present during the training phase.

Most of existing semi-supervised learning algorithms focus on classification problems, which are primarily based on continuity and cluster assumptions. Continuity assumes that neighboring data points likely share the same label, clustering of the data makes it possible to draw decision boundaries in low density regions \cite{zhu2009introduction,chapelle2009semi}. Continuity and cluster assumptions cannot be leveraged in semi-supervised regression to the same extent as in semi-supervised classification. That is why it is more difficult to develop semi-supervised regression methods. This obstacle explains the scarcity of publications on semi-supervised regression and why research in semi-supervised regression was not able to achieve the same level of success as semi-supervised classification.

Semi-supervised regression was a research area before neural networks became popular in 2012 \cite{krizhevsky2012imagenet}. However, only very few research articles have since included neural networks as part of their proposed semi-supervised regression pipelines, mostly combined with other machine learning algorithms \cite{jean2018semi,liu2020metric}. While neural networks are not always the optimal choice, their performance ceiling is much higher due to the universal approximation theorem \cite{cybenko1989approximation,hornik1991approximation}, the incorporation of feature extraction into the learning problem, and the efficient training through gradient descent and backpropagation.

In this article we develop a semi-supervised training procedure for twin neural network regression (TNNR) \cite{wetzel2020twin}. TNNR is based on an architecture similar to Siamese neural networks \cite{bromley1993signature,baldi1993neural} in the sense that it takes two inputs. TNNR is trained to predict the differences between the labels of the input pair. The solutions of the original regression problem is then obtained by creating an ensemble of all predicted differences between a new data point and all labelled training data points plus their labels. TNNR is normally trained on pairs, here we explain how TNNR can be trained on triples of data points which can be unlabelled. These triples form a loop along which predictions must sum to zero. One can implement this constraint using a suitable loss function to transform TNNR into a semi-supervised regression algorithm. The strengths of TNNR can be summarized as follows:

\begin{enumerate}
\item TNNR attempts to circumvent the bias-variance tradeoff by internal ensembling which translates to a smaller expected error \cite{wetzel2020twin}.
\item TNNR provides uncertainty estimates through the variance of predictions \cite{wetzel2020twin}.
\item TNNR can be trained in a semi-supervised manner on loops containing unlabelled data points [This work].
\end{enumerate}

\section{Prior Work}

Twin neural networks are inspired by Siamese neural networks, these networks were introduced to solve an infinite class classification problem as it occurs in finger print recognition or signature verification \cite{bromley1993signature,baldi1993neural}. Siamese neural networks consist of two identical neural networks which project an input pair into a latent space. The similarity of two inputs is determined based on the distance in latent space. The twin neural network in TNNR also takes a pair of inputs to predict the difference between the labels \cite{wetzel2020twin}. These networks differ from Siamese networks in two properties: First, there is no (or only minimal) weight sharing between neurons acting on similar parts of each member of the input pair. Second, they do not project into a latent space but are fully connected. TNNR predictions can be transformed to a solution of the original regression problem via averaging over all predicted differences between an unlabelled data point and all labelled training data points, plus their labels. TNNR attempts to circumvent the bias-variance tradeoff, since the bias of neural networks is low given sufficiently many parameters and the variance can be reduced by the implicit ensemble of different predictions. TNNR also provides prediction uncertainty estimates through the variance of the predictions.

Most previous research in semi-supervised regression can be classified into three categories: Co-training, kernel and graph based regression methods \cite{kostopoulos2018semi}. Co-training denotes alternating training between two different regressors (the original idea is to use different feature sets for each regressor) where the predictions of one helps create a training set for the other, this process is repeated until convergence \cite{blum1998combining}. While initially invented for semi-supervised classification, there has been progress to adapt co-training to semi-supervised regression \cite{zhou2005semi,wang2010new,hady2009semi}. Semi-supervised support vector machines \cite{bennett1999semi,chapelle2008optimization} have been extended for semi-supervised regression \cite{xu2011semi}. In graph based semi-supervised regression methods labelled and unlabelled data are considered nodes on a graph. Similar nodes are connected with edges along which information propagates \cite{zhu2003semi,zhu2006semi,timilsina2021semi,timilsina2021semi}. 


\section{Twin Neural Network Regression}
\begin{figure}
    \centering
    \includegraphics[width=\textwidth]{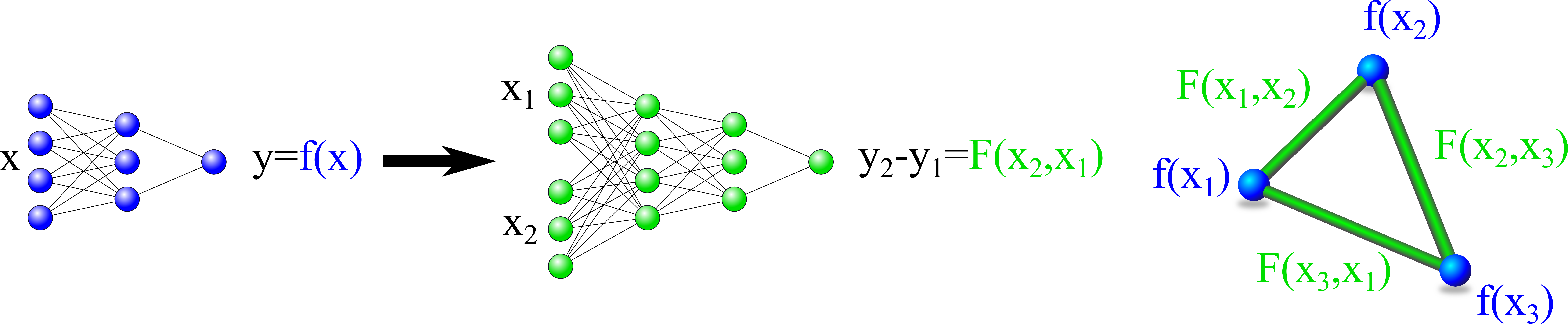}
    \caption{TNNR formulation of a regression problem: In order to solve a traditional regression problem, a neural network is trained to map a data point $x$ to its target value $f(x)=y$. A TNN takes two inputs $x_1$ and $x_2$ and is trained to predict the difference between the target values $F(x_2,x_1)=y_2-y_1$. This difference can be employed as an estimator for $y_2=F(x_2,x_1)+y_1$ given a labelled anchor point $(x_1,y_1)$. In contrast to a traditional neural network $f$, a TNN $F$ must satisfy loop consistency, predictions along each loop sum to zero:  $F(x_1,x_2)+F(x_2,x_3)+F(x_3,x_1)=0$.}
    \label{fig:architecture}
\end{figure}
\subsection{Reformulation of the Regression Problem}
We are given a training set of $n$ labelled data points $X^{train}=(x_1^{train},...,x_n^{train})$ with target values $Y^{train}=(y_1^{train},...,y_m^{train})$. The goal is to find a function $f$ such that $f(x_i)= y_i$, which generalizes to unseen data $X^{test}$ to make an accurate prediction for the labels $Y^{test}$. TNNR aims to solve a reformulation of the original regression problem, see \fig{fig:architecture}. Given a pair of data points $(x_i^{train},x_j^{train})$ we train a neural network to find a function $F$ to predict the difference 
\begin{align}
    F(x_i,x_j)= y_i-y_j \quad .    \label{eq:TNNR}
\end{align}
The TNN $F$ provides a solution to the original regression problem of predicting $y_i$ by evaluating $y_i= F(x_i,x_j)+y_j$, where $(x_j,y_j)$ is an {\em anchor} whose label is known. Every training data point $x_j^{train}\in X^{train}$ can be employed as anchor. Since ensembles of predictions are more accurate than single predictions, the best estimate for the solution of the original regression problem is obtained by averaging
\begin{align}
    y_i^{pred}&=  \frac{1}{m}\sum_{j=1}^m F(x_i,x_j^{train})+y_j^{train}= \frac{1}{m}\sum_{j=1}^m \frac{1}{2}F(x_i,x_j^{train})-\frac{1}{2}F(x_j^{train},x_i)+y_j^{train}  \quad .\label{eq01}
\end{align}

This ensemble contains twice the size of the training set of predictions of differences $y_i-y_j$ for a single prediction of $y_i$. 

Since training data points are not just separated by infinitesimal perturbations the created ensemble is more diverse that a pseudo ensemble generated through infinitesimal perturbations of model weights \cite{bachman2014learning}. Ensembles of different TNN models, each containing an implicit ensemble of predictions themselves is even more accurate than an implicit ensemble from a single TNN \cite{wetzel2020twin}.

\subsection{Semi-Supervised Learning on Loops}
\begin{figure}
    \centering
    \includegraphics[width=\textwidth]{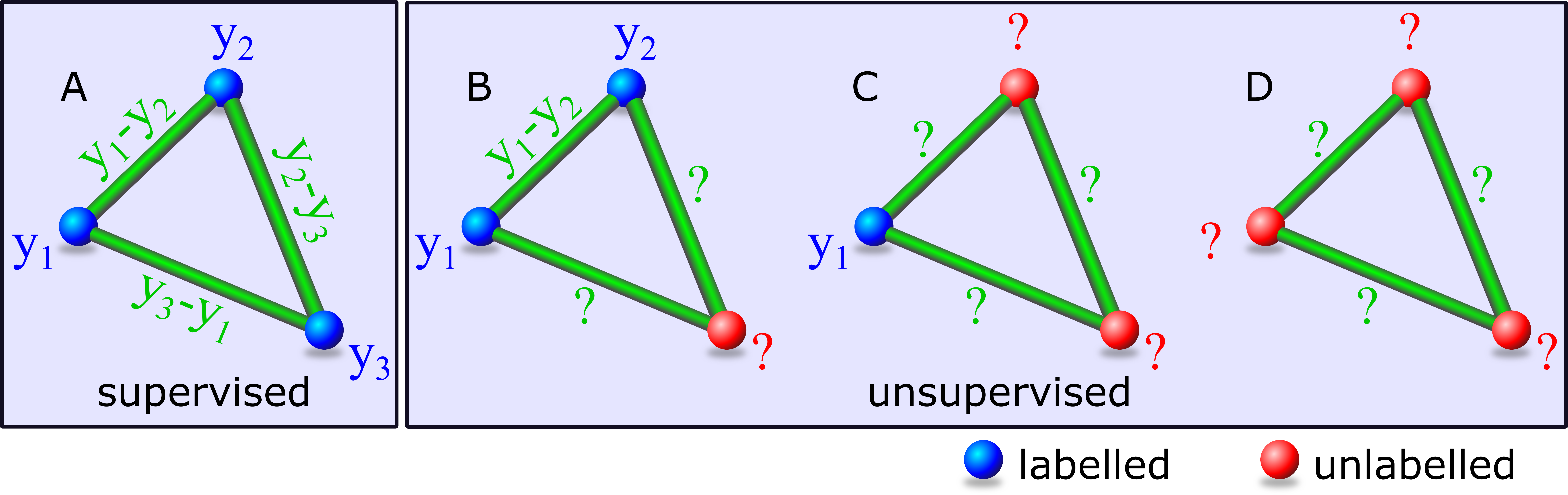}
    \caption{Training Data: A triple of labelled data points $(x_1,x_2,x_3)$ with labels $(y_1,y_2,y_3)$ translates to a TNNR training loop which has labels $(y_1-y_2, y_2-y_3, y_3-y_1)$, depicted as loop A. Loop A is used as supervised training data for TNNR with the objective to minimize the MSE between the TNN outputs and each label of loop A. In loops B, C and D at least one data point is unlabelled. On these loops, TNNR is trained in unsupervised manner to enforce loop consistency $F(x_1,x_2)+F(x_2,x_3)+F(x_3,x_1)=0$.}
    \label{fig:loops}
\end{figure}

If we have in addition to a labelled training data set of size $m$ an unlabelled data set of size $n$ available during the training phase, TNNR can be used as a semi-supervised regression method. TNNR can be trained on unlabelled data points which are connected along loops, see \fig{fig:loops}. Loops consist of several data points which can be labelled or unlabelled, the corresponding loop labels express the differences between the target labels of each original data point. For simplicity, we restrict ourselves to loops of size three. Since all higher order loops can be combined from loops of size three, we do not believe they would lead to an improved performance. In our case, a loop can be expressed as triple $(x_i,x_j,x_k)$ with data labels $(y_i,y_j,y_k)$ and corresponding loop labels $(y_i-y_j, y_j-y_k, y_k-y_i)$, if a label of a data point is unknown during the training phase the corresponding loop labels are left blank. As long as the labels are known, each pair of original data points $(x_i,x_j)$ drawn from the triple can be used to train the TNN to predict the difference $y_i-y_k$.
Furthermore, predictions on along each loop containing $(x_i,x_j,x_k)$ points should satisfy the loop consistency condition
\begin{align}
0=F(x_i,x_j)+F(x_j,x_k)+F(x_k,x_i) \quad . \label{eq:loop}
\end{align}
This condition can be used in order to train the TNN in an unsupervised manner since it does not require the presence of any labels. As shown in \fig{fig:loops} we consider four types of loops. We only use loops where all $x_i$ have a label $y_i$ as supervised training data (loop A). If at least one of the $x_i$ is unlabelled (loop B, C, D) we choose to use the loop as unsupervised training data. TNNR cannot be used as a purely unsupervised regression algorithm, some labelled data points must be contained in the training set.

\begin{figure}
    \centering
    \includegraphics[width=\textwidth]{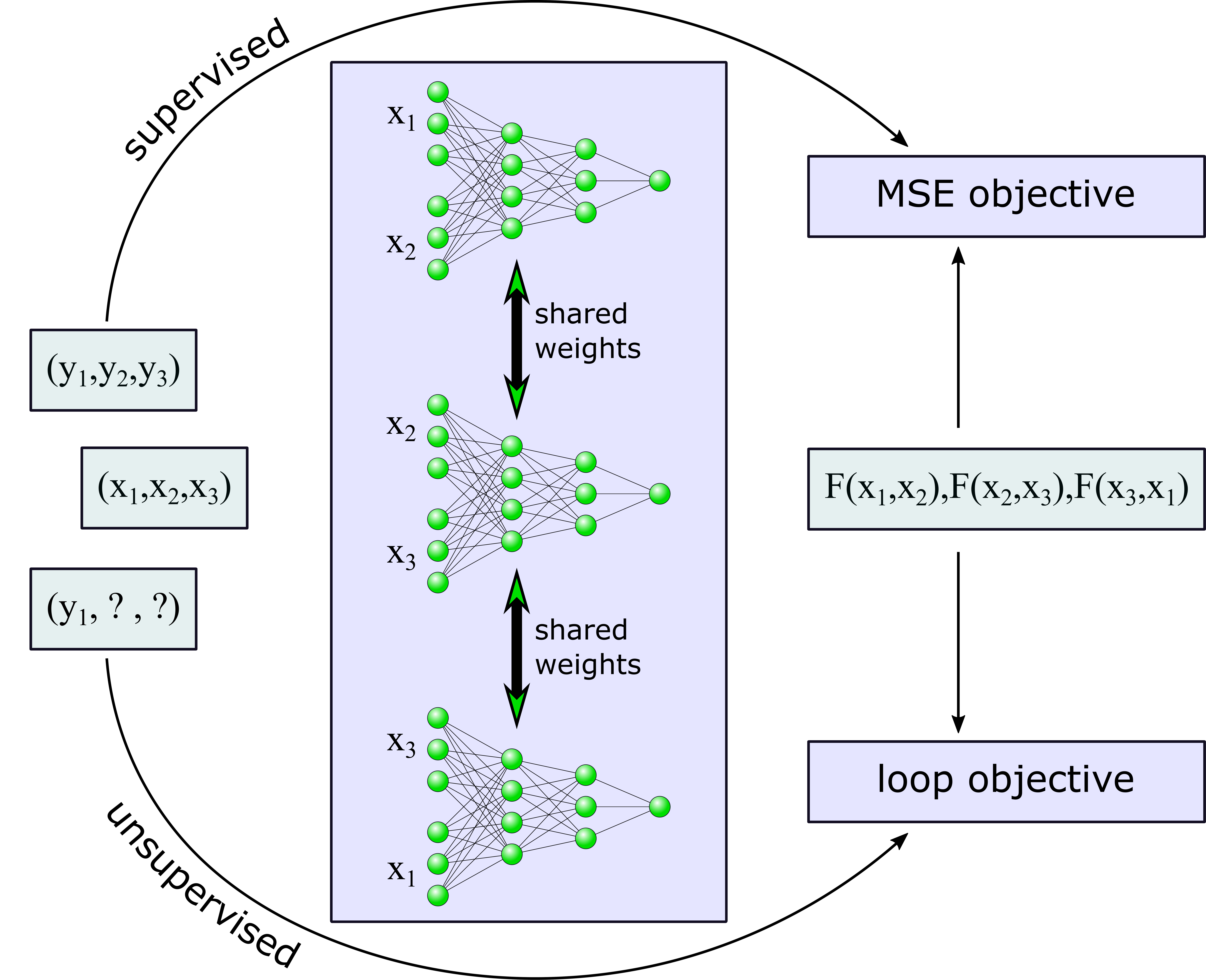}
    \caption{Training Architecture: In order to train TNNR in a semi-supervised manner, three copies of the main TNN are embedded in the training architecture. The inputs are loops, meaning triples of data points $(x_1,x_2,x_3)$ which are distributed among the three TNNs to predict differences between the target values $(y_1-y_2,y_2-y_3,y_3-y_1)$. If training data is labelled, the three TNNs are optimized through the MSE objective; if the training data is unlabelled or partially labelled, the three TNNs are optimized through the loop objective. TNNs are trained simultaneously on both labelled and unlabelled/partially labelled loops contained in the same training batch.}
    \label{fig:train}
\end{figure}

\subsection{Training Architecture}
\label{sec:training_architecture}
The architecture used to train the TNN is shown in \fig{fig:train}. Three copies of the main TNN are trained simultaneously. During training these networks share their weights. The input of the full architecture are loops along triples $(x_i,x_j,x_k)$, which can either be fully labelled or unlabelled/partially labelled. The input of the three TNNs are the three possible pairs drawn from the loop, so that the prediction of the full training model is a triple $(F(x_i,x_j),F(x_j,x_k),F(x_k,x_i))$.
Depending on the loop type \fig{fig:loops} of the input, labelled loops (loop A) are used in the mean squared error loss function (MSE objective) to train the TNN to predict the differences between each data label $(y_i-y_j, y_j-y_k, y_k-y_i)$. Unlabelled loops (loops B, C, D) are used to train the TNN to enforce loop consistency \eq{eq:loop}. For this purpose the weights and biases in the TNNs are updated via gradient descent by minimizing the batchwise estimate of 
\begin{align}
loss_{MSE}=\frac{1}{n^2}\sum_{ij}\left( F(x_i,x_j)-(y_i-y_j) \right)^2
\end{align}
if the loops are labelled and 
\begin{align}
loss_{loop}=\frac{1}{(m+n)^3}\sum_{ijk}\left( F(x_i,x_j)+F(x_j,x_k)+F(x_k,x_i)) \right)^2
\end{align}
if the loops are unlabelled or partially labelled. The loop loss function can be seen as a MSE objective with noisy labels for each of the $F(x_i,x_j)$ provided by the other two predictions $y_i-y_j\approx-F(x_j,x_k)-F(x_k,x_i)$. The combined loss function is 
\begin{align}
loss=loss_{MSE}+\Lambda \ loss_{loop} \label{eq:full_loss}
\end{align}
where the optimal loop weight $\Lambda$ is a hyperparameter that needs to be optimized by evaluating the loss on a validation set.

Our TNNs all consist of two hidden layers, each with 192 neurons in these layers. Each hidden layer uses rectified linear units as activation functions while the output is a linear neuron. We do not employ any regularization. We note, it is very inconvenient to store all possible loop combinations as input data for the training architecture. For this reason, we employ a generator which randomly samples all possible loops containing all original data points, labelled and unlabelled. Each batch generated by the generator contains the same number of supervised loops and unsupervised loops. The training batch size is set to 16, where half of the batch is used for supervised learning and the other half for unsupervised learning. We train for 2000 epochs using the adadelta optimizer. We employ learning rate decay callbacks and early stopping callbacks stop the training if the validation loss stops improving.

For labelled and unlabelled data set sizes of $m$ and $n$, respectively, the time complexity estimate for traditional neural networks scales like $\mathcal{O}(m)$ for training and $\mathcal{O}(1)$ for inference. This translates to a training time complexity for TNNs of $\mathcal{O}((m+n)^2)$, since the number of possible differences scales quadratically with the number original data points. Since the loop loss function behaves similar to the MSE loss function with noisy labels, we expect quadratic scaling in both  supervised and semi-supervised training. Ensembling the predictions for the differences between all labelled training data points and a new data point during inference time scales like $\mathcal{O}(m)$. For training and inference only a subset of all possible differences can be sampled to reduce time at the cost of accuracy. 

Let us discuss if it is possible to relate TNNR to the existing classes of semi-supervised regression methods which are co-training, kernel or graph based \cite{kostopoulos2018semi}. TNNR cannot be classified into either of these classes, however some ideas might appear similar. In a vague way, it is possible to argue that TNNR is similar to co-training since in the loop loss function two predictions act as a noisy label for the third prediction. Kernel methods make use of some sort of similarity kernel function, TNNR provides a similarity measure in the label space through \eq{eq:TNNR}. Further, TNNR can be connected to graph based methods, since the TNNR training loops are minimal sub graphs containing three nodes and three edges.

\section{Experiments}
\begin{figure}
    \centering
    \includegraphics[width=\textwidth]{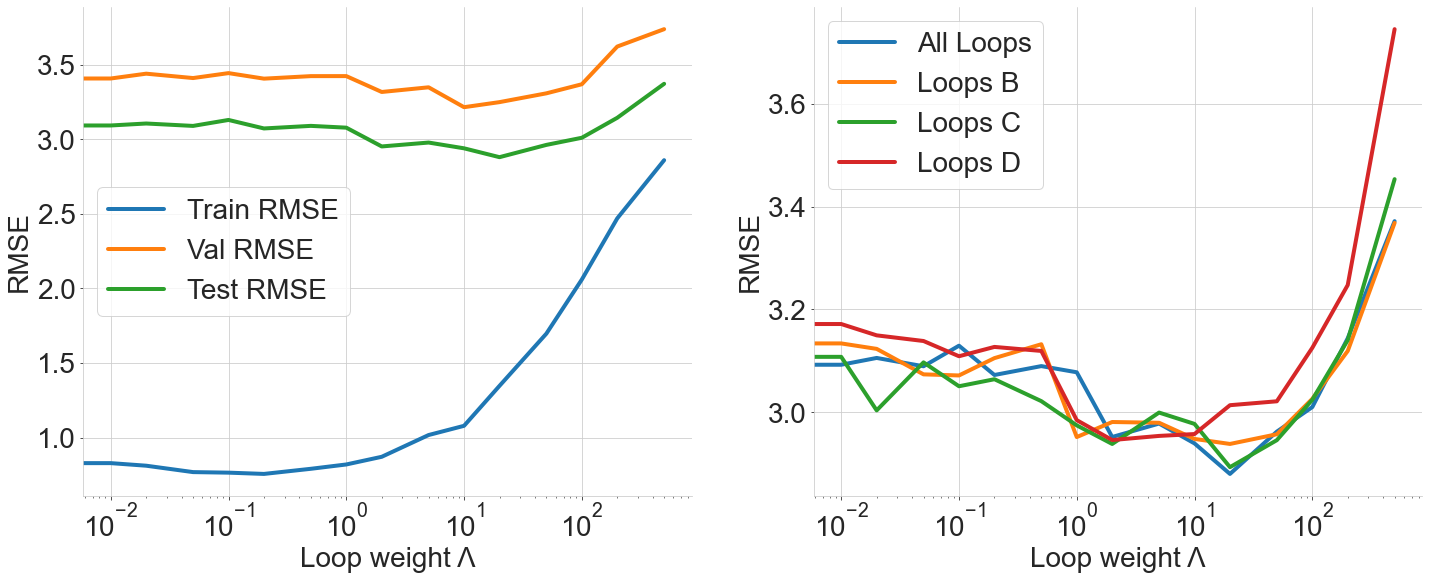}
    \caption{Effect of semi-supervised training on RMSE at the example of the Boston Housing data set. The loop weight denotes the strength of the loop constraint versus the MSE loss function, see \eq{eq:full_loss}. Left: Increasing the loop weight $\Lambda$ to a suitable value causes the validation and test RMSE to decrease while at the same time the train RMSE increases. Right: different loop types reduce the RMSE in a similar manner, All loops together perform slightly better.}
    \label{fig:loop_comparison}
\end{figure}

\subsection{Data Preparation}
\label{sec:data}
We examine the performance of TNNR on different regression data sets: bio conservation (BC), boston housing (BH), concrete strength (CS), energy efficiency (EE), RCL circuit (RCL), red wine quality (WN), test function (TF), red wine quality (WN), Wheatstone bridge (WSB) and yacht hydrodynamics (YH). The common data sets can be found online at \cite{datasets}. The scientific data sets are simulations of mathematical equations and physical systems. TF is a polynomial function combined with a sin function. RCL is a simulation of the electric current in an RCL circuit and WSB a simulation of the Wheatstone bridge voltage. More details can be found in the supplementary material.

We perform two different experiments to examine TNNR as transductive and inductive semi-supervised learning algorithm. To focus solely on transductive learning, in the first experiments all data is split into 80\% labelled training, 10\% unlabelled validation and 10\% unlabelled test data, all of which is used during the training phase. To examine inductive learning in addition to transductive learning we perform a second experiment with 30\% labelled training, 50\% unlabelled validation and 10\% unlabelled test data, which is used during training and 10\% unlabelled data on which the performance of inductive semi-supervised learning is evaluated. We normalize and center the input features to a range of $[-1,1]$ based on the training data. As outlined in \sec{sec:training_architecture}, labelled and unlabelled data are combined to form loops, which are supplied to the training architecture \fig{fig:train} through a generator from which all possible loops are sampled. All resulting RMSEs in the figures and tables are produced by repeating the experiment 25 times which random splits of the underlying data, the same random splits are used while varying the hyperparameter $\Lambda$.

\subsection{Experiment 1: Transductive Semi-Supervised Learning}
\begin{table}
\centering
  \caption{Best estimates for test RMSEs belonging to different data sets. Our confidence on the RMSEs is determined by their standard error. Data sets: bio conservation (BC), boston housing (BH), concrete strength (CS), energy efficiency (EE), RCL circuit (RCL), red wine quality (WN), test function (TF), red wine quality (WN), Wheatstone bridge (WSB) and yacht hydrodynamics (YH). We train on 100\% of the available data where 80\% is labelled training data, 10\% is unlabelled validation data and 10\% is unlabelled test data whose labels are predicted using TNNR as a transductive semi-supervised learning method.
  }
 
  \begin{tabular}{l llr}
    & \multicolumn{3}{l}{80\% labelled training data}            \\
    \cmidrule(r){2-2}\cmidrule(r){3-4}
    &Supervised&Transductive&Gain\\
    BC &$0.8121\!\pm\!0.0219$&$0.7221\!\pm\!0.0178$&$11.1\%$\\
    BH &$3.0716\!\pm\!0.1678$&$2.9205\!\pm\!0.1642$&$4.9\%$\\
    CS &$4.3043\!\pm\!0.1476$&$4.3043\!\pm\!0.1476$&$0.0\%$\\
    EE &$0.6143\!\pm\!0.0236$&$0.5647\!\pm\!0.0179$&$8.1\%$\\
    RCL &$0.0145\!\pm\!0.0002$&$0.0143\!\pm\!0.0002$&$1.3\%$\\
    TF &$0.0044\!\pm\!0.0002$&$0.0038\!\pm\!0.0002$&$13.6\%$\\
    WN &$0.7037\!\pm\!0.0100$&$0.6334\!\pm\!0.0084$&$10.0\%$\\
    WSB &$0.0214\!\pm\!0.0011$&$0.0214\!\pm\!0.0011$&$0.0\%$\\
    YH &$0.5597\!\pm\!0.0592$&$0.5597\!\pm\!0.0592$&$0.0\%$\\

  \end{tabular}
  \label{tab:results1}
\end{table}

The first experiment focuses on transductive semi supervised learning, where the goal is to predict labels of unlabelled data which is present during the training phase. First, for the example of the boston housing data set, we include all possible loop types depicted in \fig{fig:loops} into the training data and examine if the effect on the RMSEs of the training, validation and test sets. In \fig{fig:loop_comparison}, we can see that the validation and test set RMSE can be lowered by tuning the loop weight $\Lambda$ to a certain finite value. However, at that point the training RMSE increases above its base value. This indicates that including the loops containing unlabelled data points is similar to regularization and reduces overfitting to the labelled training set. Secondly, we determine the effects of including several loop types into the training phase at the example of the boston housing data set. In \fig{fig:loop_comparison} we can see that all loops have a positive effect on the RMSE, however, some loops seem to have a slightly better effect than others. Loops B and C tend to reduce the RMSE more than Loop D. This is most likely because Loops B and C contain the final predictions of differences between elements of the labelled training set and the unlabelled test set. Based on this figure we choose to train on all possible loops in all further experiments.

The results of semi-supervised transductive learning for different data sets are presented in \tab{tab:results1} where they are compared with purely supervised results. Supervised learning is equivalent to setting the loop weight $\Lambda=0$. The transductive results are produced choosing the optimal loop weight $\Lambda\in [0.01,0.02,0.05,0.1,0.2,0.5,1,2,5,10,20,50,100,200,500]$ based on the validation RMSE. The RMSE curves are visualized in the supplementary material. Due to noise, sometimes the optimal choice of $\Lambda$ based on the validation RMSE slightly differs from the optimum for the test RMSE. In some cases (CS,WSB,YH) this prevents us from uniquely identifying an optimal finite $\Lambda$. In these cases we cannot claim an improvement of semi-supervised learning over purely supervised learning. In some data sets the RMSE shows two local minima as a function of $\Lambda$, this effect is separate from noise and can be reproduced, but we do not know of the cause of this observation. 

On 6 out of 9 data sets the test RMSE can be significantly reduced by semi-supervised transductive learning. An interesting observation is that the data sets which did not benefit from internal TNNR ensembling in the original TNNR paper \cite{wetzel2020twin}, namely bio conservation and red wine quality, are among those that benefit most from semi-supervised learning. That means TNNR ensembling combined with semi-supervised training on loops improved the RMSE on all considered data sets significantly compared to previous state-of-the art regression methods.

\subsection{Experiment 2: Transductive and Inductive Semi-Supervised Learning}
In the second experiment, the goal is to compare transductive and inductive semi-supervised learning using TNNR. The transductive RMSEs are evaluated on the unlabelled test data which was supplied to the training phase, while the inductive RMSEs are evaluated on the the unlabelled test set which was not contained in the training loops. Further, the validation set was part of the training loops. In \tab{tab:results2} the improvements due to semi-supervised training are displayed, while the RMSE curves can be found in the supplementary material. From this table we conclude that TNNR can be successfully used as a transductive or inductive semi-supervised learning method. 
\begin{table}
\centering
  \caption{Best estimates for test RMSEs belonging to different data sets. Our confidence on the RMSEs is determined by their standard error. Data sets: bio conservation (BC), boston housing (BH), concrete strength (CS), energy efficiency (EE), RCL circuit (RCL), red wine quality (WN), test function (TF), red wine quality (WN), Wheatstone bridge (WSB) and yacht hydrodynamics (YH). We train on 90\% of the available data where 30\% is labelled training data, 10\% is unlabelled validation data and 50\% is unlabelled test data whose labels are predicted using TNNR as a transductive semi-supervised learning method. The labels of the 10\% of the data which was not used during training are inferred using TNNR as an inductive semi-supervised learning method.
  }
 
  \resizebox{\textwidth}{!}{
  \begin{tabular}{l llr llr}
    & \multicolumn{3}{l}{30\% labelled training data}            \\
    \cmidrule(r){2-4}\cmidrule(r){5-7}
    &Supervised&Transductive&Gain&Supervised&Inductive&Gain\\

    BC &$0.9382\!\pm\!0.0137$&$0.7960\!\pm\!0.0056$&$15.2\%$&$0.8996\!\pm\!0.0280$&$0.7721\!\pm\!0.0175$&$14.2\%$\\
    BH &$4.1357\!\pm\!0.1229$&$3.8228\!\pm\!0.0951$&$7.6\%$&$3.6830\!\pm\!0.2337$&$3.5521\!\pm\!0.2281$&$3.6\%$\\
    CS &$6.0777\!\pm\!0.0773$&$5.9088\!\pm\!0.0616$&$2.8\%$&$6.0467\!\pm\!0.1412$&$6.0905\!\pm\!0.1260$&$-1.0\%$\\
    EE &$1.5084\!\pm\!0.0317$&$1.4194\!\pm\!0.0409$&$5.9\%$&$1.4794\!\pm\!0.0416$&$1.3902\!\pm\!0.0459$&$6.0\%$\\
    RCL &$0.0200\!\pm\!0.0003$&$0.0194\!\pm\!0.0003$&$3.0\%$&$0.0203\!\pm\!0.0004$&$0.0195\!\pm\!0.0004$&$3.9\%$\\
    TF &$0.0066\!\pm\!0.0004$&$0.0063\!\pm\!0.0004$&$4.5\%$&$0.0064\!\pm\!0.0004$&$0.0059\!\pm\!0.0004$&$7.8\%$\\
    WN &$0.7841\!\pm\!0.0047$&$0.6511\!\pm\!0.0027$&$17.0\%$&$0.7868\!\pm\!0.0087$&$0.6534\!\pm\!0.0075$&$17.0\%$\\
    WSB &$0.0341\!\pm\!0.0012$&$0.0341\!\pm\!0.0012$&$0.0\%$&$0.0368\!\pm\!0.0018$&$0.0368\!\pm\!0.0018$&$0.0\%$\\
    YH &$1.2203\!\pm\!0.0616$&$1.2203\!\pm\!0.0616$&$0.0\%$&$1.1170\!\pm\!0.0910$&$1.1170\!\pm\!0.0910$&$0.0\%$\\
  
  \end{tabular}
  }
  \label{tab:results2}
\end{table}

\section{Summary}

We have presented a method to train twin neural network regression (TNNR) in a semi-supervised manner. While TNNR is already a state of the art regression method, we observe a significant improvement by semi supervised training on loops containing differences between labelled and/or unlabelled data. On 6 out of 9 data sets the test RMSE can be significantly reduced by semi-supervised learning. Data sets which do not benefit from internal ensembling \cite{wetzel2020twin} benefit among the most from semi-supervised learning. Thus, TNNR ensembling combined with semi-supervised training on loops beat all considered supervised state-of-the-art methods on all considered data sets. This outperformance is persistent no matter if it is achieved via transductive or inductive semi-supervised learning.
The effective cost of TNNR compared to traditional supervised neural networks is the quadratic time scaling of the former versus the linear time scaling of the latter.

In the future, it might be helpful to compare different semi-supervised regression methods to each other. It might be possible to improve the results by continuously increasing the loop weight $\Lambda$ during training. It is very likely that certain specific loops improve the performance more than others, after identifying these loops the learning process could be weighted in their favor. While in this work only the training process is assisted by loops, it might be fruitful to explore if the loops can help during the inference phase.

\section{Acknowledgements}
Research at Perimeter Institute is supported in part by the Government of Canada through the Department of Innovation, Science and Economic Development Canada and by the Province of Ontario through the Ministry of Economic Development, Job Creation and Trade. We thank the National Research Council of Canada for their partnership with Perimeter on the PIQuIL. R.G.M. and I.T. acknowledge NSERC. R.G.M. is supported by the Canada Research Chair Program. We also acknowledge Compute Canada for computational resources.

\medskip

\small

\bibliographystyle{plain}
\bibliography{library}

\begin{thebibliography}{10}

\bibitem{bachman2014learning}
Philip Bachman, Ouais Alsharif, and Doina Precup.
\newblock Learning with pseudo-ensembles.
\newblock {\em arXiv preprint arXiv:1412.4864}, 2014.

\bibitem{baldi1993neural}
Pierre Baldi and Yves Chauvin.
\newblock Neural networks for fingerprint recognition.
\newblock {\em neural computation}, 5(3):402--418, 1993.

\bibitem{bennett1999semi}
Kristin Bennett, Ayhan Demiriz, et~al.
\newblock Semi-supervised support vector machines.
\newblock {\em Advances in Neural Information processing systems}, pages
  368--374, 1999.

\bibitem{blum1998combining}
Avrim Blum and Tom Mitchell.
\newblock Combining labeled and unlabeled data with co-training.
\newblock In {\em Proceedings of the eleventh annual conference on
  Computational learning theory}, pages 92--100, 1998.

\bibitem{bromley1993signature}
Jane Bromley, Isabelle Guyon, Yann LeCun, Eduard S{\"a}ckinger, and Roopak
  Shah.
\newblock Signature verification using a" siamese" time delay neural network.
\newblock {\em Advances in neural information processing systems}, 6:737--744,
  1993.

\bibitem{chapelle2009semi}
Olivier Chapelle, Bernhard Scholkopf, and Alexander Zien.
\newblock Semi-supervised learning.
\newblock {\em IEEE Transactions on Neural Networks}, 20(3):542--542, 2009.

\bibitem{chapelle2008optimization}
Olivier Chapelle, Vikas Sindhwani, and Sathiya~S Keerthi.
\newblock Optimization techniques for semi-supervised support vector machines.
\newblock {\em Journal of Machine Learning Research}, 9(2), 2008.

\bibitem{cybenko1989approximation}
George Cybenko.
\newblock Approximation by superpositions of a sigmoidal function.
\newblock {\em Mathematics of control, signals and systems}, 2(4):303--314,
  1989.

\bibitem{hady2009semi}
Mohamed Farouk~Abdel Hady, Friedhelm Schwenker, and G{\"u}nther Palm.
\newblock Semi-supervised learning for regression with co-training by
  committee.
\newblock In {\em International Conference on Artificial Neural Networks},
  pages 121--130. Springer, 2009.

\bibitem{hornik1991approximation}
Kurt Hornik.
\newblock Approximation capabilities of multilayer feedforward networks.
\newblock {\em Neural networks}, 4(2):251--257, 1991.

\bibitem{jean2018semi}
Neal Jean, Sang~Michael Xie, and Stefano Ermon.
\newblock Semi-supervised deep kernel learning: Regression with unlabeled data
  by minimizing predictive variance.
\newblock {\em arXiv preprint arXiv:1805.10407}, 2018.

\bibitem{kostopoulos2018semi}
Georgios Kostopoulos, Stamatis Karlos, Sotiris Kotsiantis, and Omiros Ragos.
\newblock Semi-supervised regression: A recent review.
\newblock {\em Journal of Intelligent \& Fuzzy Systems}, 35(2):1483--1500,
  2018.

\bibitem{krizhevsky2012imagenet}
Alex Krizhevsky, Ilya Sutskever, and Geoffrey~E Hinton.
\newblock Imagenet classification with deep convolutional neural networks.
\newblock {\em Advances in neural information processing systems},
  25:1097--1105, 2012.

\bibitem{liu2020metric}
Chien-Liang Liu and Qing-Hong Chen.
\newblock Metric-based semi-supervised regression.
\newblock {\em IEEE Access}, 8:30001--30011, 2020.

\bibitem{timilsina2021semi}
Mohan Timilsina, Alejandro Figueroa, Mathieu d’Aquin, and Haixuan Yang.
\newblock Semi-supervised regression using diffusion on graphs.
\newblock {\em Applied Soft Computing}, 104:107188, 2021.

\bibitem{datasets}
UCI.
\newblock Machine learning repository
  https://archive.ics.uci.edu/ml/datasets.php (accessed: 1 may 2020).

\bibitem{wang2010new}
Wei Wang and Zhi-Hua Zhou.
\newblock A new analysis of co-training.
\newblock In {\em ICML}, 2010.

\bibitem{wetzel2020twin}
Sebastian~J Wetzel, Kevin Ryczko, Roger~G Melko, and Isaac Tamblyn.
\newblock Twin neural network regression.
\newblock {\em arXiv preprint arXiv:2012.14873}, 2020.

\bibitem{xu2011semi}
Shuo Xu, Xin An, Xiaodong Qiao, Lijun Zhu, and Lin Li.
\newblock Semi-supervised least-squares support vector regression machines.
\newblock {\em journal of information \& computational science}, 8(6):885--892,
  2011.

\bibitem{zhou2005semi}
Zhi-Hua Zhou and Ming Li.
\newblock Semi-supervised regression with co-training.
\newblock In {\em IJCAI}, volume~5, pages 908--913, 2005.

\bibitem{zhu2003semi}
Xiaojin Zhu, Zoubin Ghahramani, and John~D Lafferty.
\newblock Semi-supervised learning using gaussian fields and harmonic
  functions.
\newblock In {\em Proceedings of the 20th International conference on Machine
  learning (ICML-03)}, pages 912--919, 2003.

\bibitem{zhu2006semi}
Xiaojin Zhu and Andrew Goldberg.
\newblock Semi-supervised regression with order preferences.
\newblock Technical report, University of Wisconsin-Madison Department of
  Computer Sciences, 2006.

\bibitem{zhu2009introduction}
Xiaojin Zhu and Andrew~B Goldberg.
\newblock Introduction to semi-supervised learning.
\newblock {\em Synthesis lectures on artificial intelligence and machine
  learning}, 3(1):1--130, 2009.

\end{thebibliography}

\newpage
\section{Supplementary Material}

\begin{table}[h!]
  \caption{Datasets
  }
  \centering
  \begin{tabular}{lllll}

  Name & Key & Size & Features & Type\\
\midrule
  Bio Concentration & BC &779&14&Discrete, Continuous\\
  Boston Housing & BH &506 & 13&Discrete, Continuous\\
  Concrete Strength & CS &1030&8&Continuous\\
  Energy Efficiency & EF &768&8& Discrete, Continuous \\
  RCL Circuit Current &RCL&4000&6&Continuous\\
  Test Function & TF & 1000 & 2 & Continuous\\
  Red Wine Quality & WN &1599&11&Discrete, Continuous\\
  Wheatstone Bridge Voltage &WSB&200&4&Continuous\\
  Yacht Hydrodynamics & YH &308&6&Discrete\\
  \end{tabular}
\end{table}
The test function (TF) data set created from the equation
\begin{align}
F(x_1,x_2)=x_1^3+x_1^2-x_1-1+x_1x_2+\sin(x_2)
\end{align}
and zero noise.

The output in the RCL circuit current data set (RCL) is the current through an RCL circuit, modeled by the equation
\begin{align}
I_0=V_0 \cos(\omega t)/\sqrt{R^2+(\omega L-1/(\omega C))^2}
\end{align}
with added Gaussian noise of mean 0 and standard deviation 0.1.

The output of the Wheatstone Bridge voltage (WSB) is the measured voltage given by the equation
\begin{align}
V=U(R_2/(R_1+R_2)-R_3/(R_2+R_3))
\end{align}
with added Gaussian noise of mean 0 and standard deviation 0.1.

\begin{figure}
    \centering
    \includegraphics[width=0.95\textwidth]{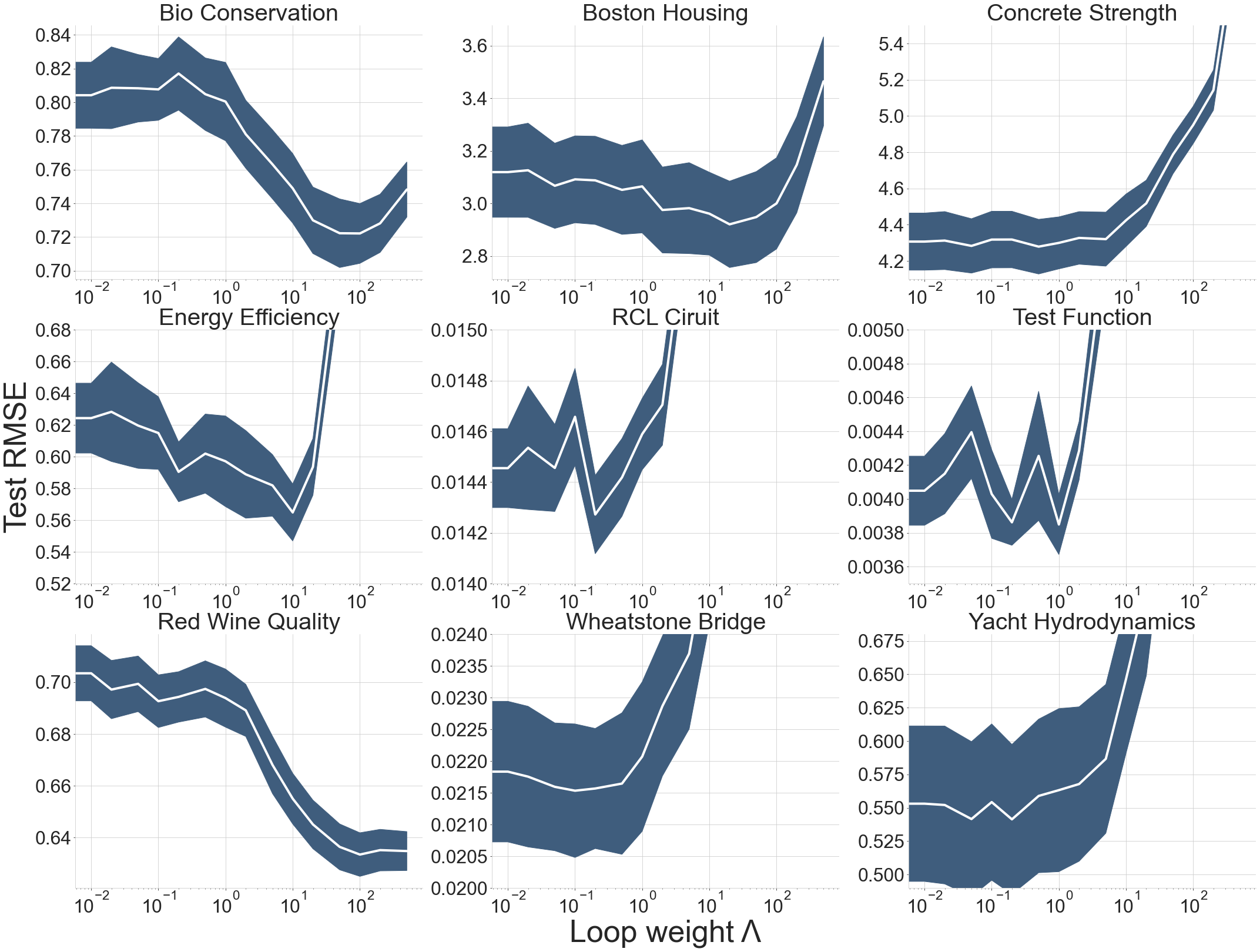}
    \caption{Effect of \textbf{transductive semi-supervised} training on RMSEs of different data sets. We train on 80\% labelled training data, 10\% unlabelled validation data and 10\% is unlabelled test data.The loop weight denotes the strength of the loop constraint versus the MSE loss function, see \eq{eq:full_loss}. The blue area is indicative for the standard error.}
    \label{fig:results1}
\end{figure}
\begin{figure}
    \centering
    \includegraphics[width=0.95\textwidth]{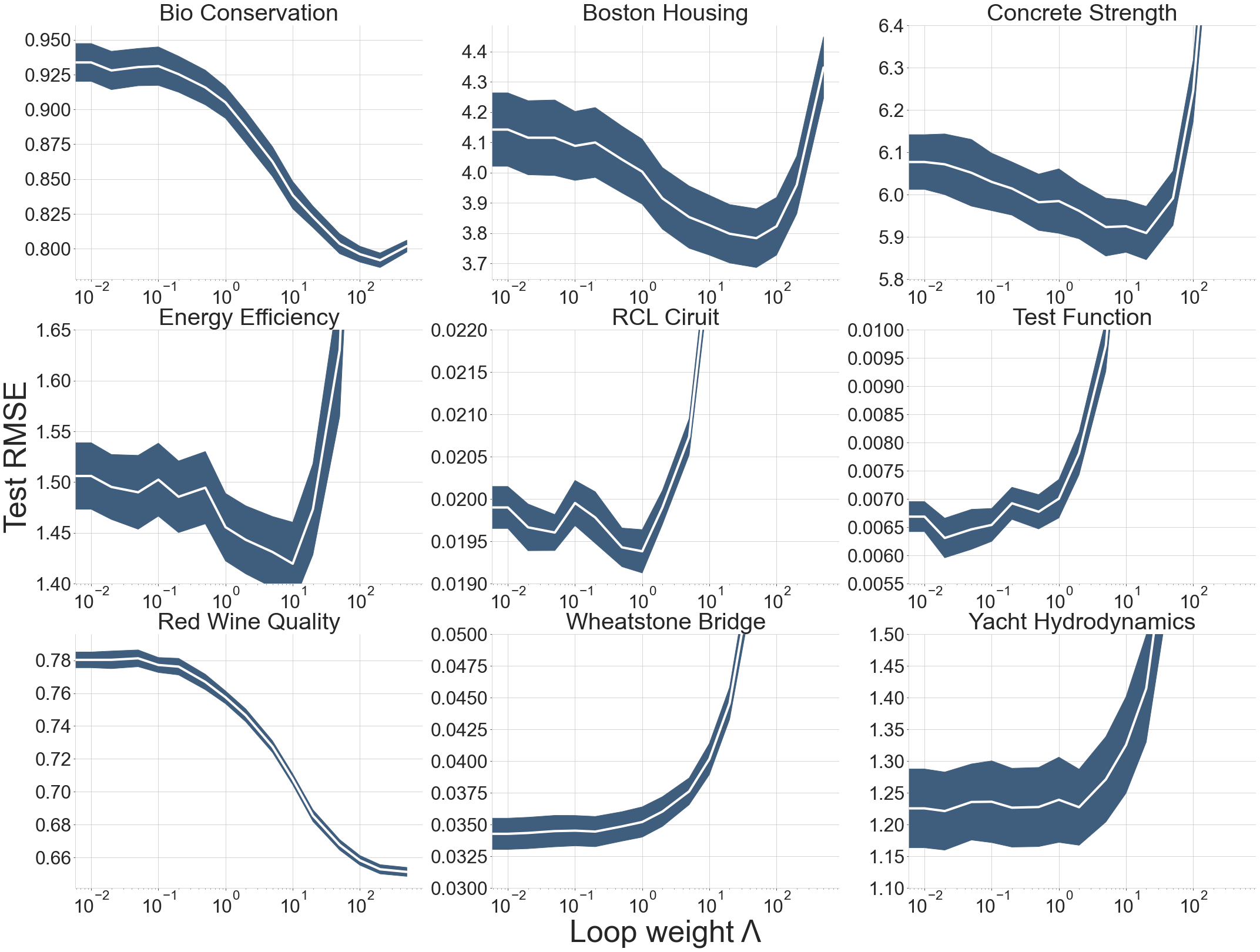}
    \caption{Effect of \textbf{transductive semi-supervised} training on RMSEs of different data sets. We train on 30\% labelled training data, 10\% unlabelled validation data and 50\% is unlabelled test data, whose targets are predicted using TNNR as a transductive semi-supervised learning method. The remaining 10\% unlabelled test data is not used for training, its labels are predicted using TNNR as an inductive semi-supervised learning method. The loop weight denotes the strength of the loop constraint versus the MSE loss function, see \eq{eq:full_loss}. The blue area is indicative for the standard error.}
    \label{fig:results2}
\end{figure}\begin{figure}
    \centering
    \includegraphics[width=0.95\textwidth]{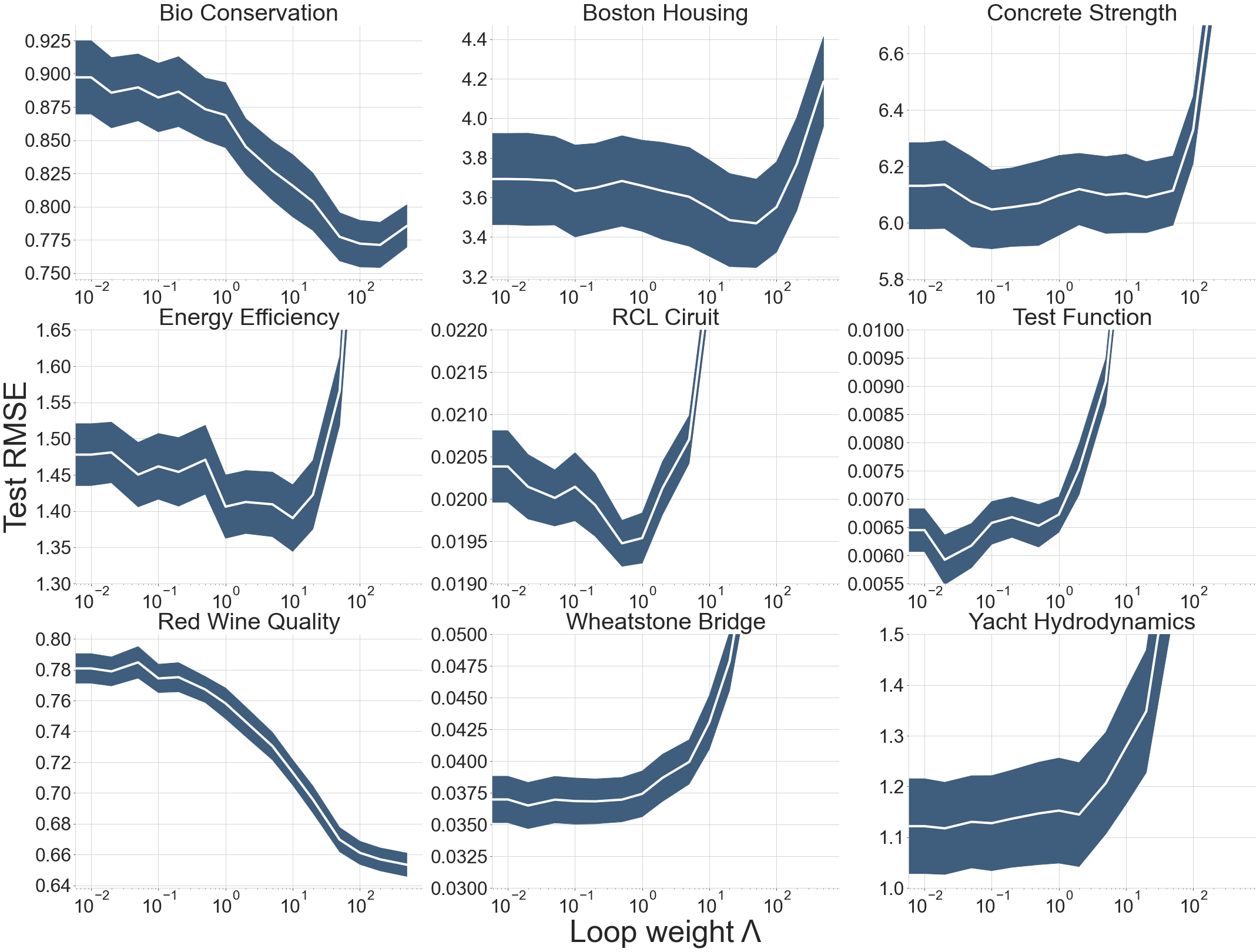}
    \caption{Effect of \textbf{inductive semi-supervised} training on RMSEs of different data sets. We train on 30\% labelled training data, 10\% unlabelled validation data and 50\% is unlabelled test data, whose targets are predicted using TNNR as a transductive semi-supervised learning method. The remaining 10\% unlabelled test data is not used for training, its labels are predicted using TNNR as an inductive semi-supervised learning method. The loop weight denotes the strength of the loop constraint versus the MSE loss function, see \eq{eq:full_loss}. The blue area is indicative for the standard error.}
    \label{fig:results3}
\end{figure}
\end{document}